\documentclass{scrartcl}

\usepackage[numbers,elide]{natbib}
\usepackage{fancybox}
\usepackage{shortvrb}
\MakeShortVerb\|
\usepackage{mathtools}
\usepackage{amsmath}
\usepackage{amssymb}
\usepackage{graphicx}
\usepackage{listings}
\usepackage{float}
\usepackage[usenames]{color}
\definecolor{codecolor}{gray}{.9}
\definecolor{rlcolor}{cmyk}{0,1,0,0}
\usepackage[multiple]{footmisc}
\usepackage{bigfoot}
\DeclareNewFootnote{AAffil}[arabic]
\DeclareNewFootnote{ANote}[fnsymbol]

\begin{document}

\title{Variational multichannel multiclass segmentation
using unsupervised lifting with CNNs
}

\author{N. Gruber \footnoteAAffil{University of Innsbruck, Technikerstra\ss e 13, 6020 Innsbruck, Austria} \ \footnoteAAffil{VASCage GmbH Anichstraße 5, 6020 Innsbruck, Austria} \and J. Schwab\footnoteAAffil{MRC-Laboratory of Molecular Biology, Francis Crick Avenue, Cambridge, UK} \and S. Court\FootnotemarkANote{1} \and E.Gizewski\FootnotemarkANote{2} \ \footnoteAAffil{Department of Neuroradiology, Medical University of Innsbruck, Austria} \and M. Haltmeier\FootnotemarkANote{1}}

\maketitle

\begin{abstract}

We propose an unsupervised image segmentation approach, that combines a variational energy functional and deep convolutional neural networks. The variational part is based on a recent multichannel multiphase Chan-Vese model, which is capable to extract useful information from multiple input images simultaneously. We implement a flexible multiclass segmentation method that divides a given image into $K$ different regions.
We use convolutional neural networks (CNNs) targeting a pre-decomposition of the image. By subsequently minimising the segmentation functional, the final segmentation is obtained in a fully unsupervised manner. Special emphasis is given to the extraction of informative feature maps serving as a starting point for the segmentation. The initial results indicate that the proposed method is able to decompose and segment the different regions of various types of images, such as texture and medical images and compare its performance with another multiphase segmentation method.
\end{abstract}

\section{INTRODUCTION}

\noindent Automating the segmentation of images plays an important role in many practical applications. In medical imaging, for example, this process is one of the most essential steps in diagnosis and treatment planning. Automating segmentation can save a lot of time and avoid subjective errors~\cite{Alaa2019CardiovascularDR,kart2021deep}.

\noindent There are at least two different classes of image segmentation methods. On the one hand, there are the classical variational image segmentation methods, where a particular energy such as the Mumford-Shah functional~\cite{Mumford1989OptimalAB} or its level set formulation proposed by Chan and Vese~\cite{chan2001active,vese2002multiphase}, is minimised. These classical models rely on the gray values of the image pixels and are therefore inadequate to capture variation in region intensities, thus resulting in improper segmentation. 

\noindent The second generation of segmentation methods consists in deep learning based models, which have superseded the classical methods in the last view years~\cite{Liu2021ARO}. These methods achieve impressive results, but have the disadvantage that most of them are supervised requiring  a large amount of hand-annotated ground truth segmentation masks.

\noindent Some works combine variational methods and deep learning by replacing operations in the minimisation algorithm of variational approaches by neural networks~\cite{boink2019partially}, while for instance in~\cite{kim2019mumford} a variational loss function is incorporated into a deep learning segmentation approach. 
In~\cite{kiechle2018model}, the authors learn linear convolution kernels followed by a single non-linearity to obtain approximately piecewise constant features for texture segmentation prior to minimisation of the Mumford-Shah model. Their approach differs from our method, as their loss function tries to minimise the jump set of the filterings and does not take into account the reconstruction of the image from the feature maps. Further, we allow for more complex CNN architectures with multiple layers, with the capability to improve the feature maps.

\noindent In this work, we address the weaknesses of classical and deep learning based segmentation methods while taking advantage of their strengths by proposing a hybrid approach that leverages the robustness of traditional classical image segmentation algorithms while exploiting the benefits of CNNs for image representation. Non-smooth level set based approaches are based on intensity values and are therefore unable to handle different types of images such as texture images. We circumvent these issues by combining the energy functional proposed in~\cite{gruber2022joint} with a CNN-based lifting strategy.
More specifically, to this end, we train CNNs to transform a single image into a multichannel representation that serves as input to a proposed multichannel extension of the Chan-Vese active contour model.

\noindent In the following section we formulate and describe our method. In section~\ref{sec3} we show some initial results of medical image segmentation, demonstrating that our method works well for this type of problems.

\section{SEGMENTATION METHOD OF~\cite{gruber2022joint}}\label{sec2}
\noindent In this section we will introduce the mathematical formulation and necessary notations of the segmentation problem, before we introduce the proposed unsupervised CNN-based lifting approach. 
In what follows, we denote by $\mathbb{F}$ the space of images modeled as functions $f:\Omega\rightarrow\mathbb{R}^d$.
Based on specific pre-defined characteristics of the image $f\in\mathbb{F}$, we want to construct a partitioning $\Omega=\bigcup_{k=0}^K\Sigma_k$ of the domain $\Omega$ into $K+1$ disjoint classes. 
\noindent Before we start describing the proposed segmentation strategy, we fix the notation. Let $BV(\Omega)$ denote the space of all integrable functions $u:\Omega\rightarrow\mathbb{R}$ with bounded total variation $\lvert u\rvert_{TV}$~\citep[Theorem~1.9, page~7]{giusti1984minimal}, where we denote $K-$tuples of functions in $BV(\Omega)$ by $\boldsymbol{u}=(u_k)_{k=1}^K$. These functions will then be used to define the partitioning of $\Omega$ as $\Sigma_k = \{x \in \Omega \mid u_k(x) > u_j(x) \text{ for all } j\neq k\}$. For these partition defining functions we define the admissible set 
\begin{align*}
    \mathbb{A}\coloneqq\{\boldsymbol{u}\in BV(\Omega)^K\mid\boldsymbol{u}\geq 0\land\sum_{k=1}^K u_k\leq 1\}.
\end{align*}
The workflow proposed in~\cite{gruber2022joint} consists of two steps, namely the input lifting, and the subsequent minimisation of the proposed energy functional. To achive the input lifting we apply $K$ feature enhancing transforms $\Phi_1,\dots,\Phi_K:\boldsymbol{F}\rightarrow L^{\infty}(\Omega)$ in a way that the resulting feature maps $\phi_k\coloneqq\Phi_k(f)$ allow to well separate region $\Sigma_k$ from $\Omega\setminus\Sigma_k$.
\noindent The second step consists of minimising the following energy functional
   \begin{align}\label{energyred}
       \mathcal{R}_{\phi,\lambda}(\mathbf{u})\coloneqq
       \begin{cases} \lambda\mathcal{Q}(\mathbf{u}) + \inf\mathcal{D}_{\phi}(\mathbf{u},\cdot) & \text{if}\quad \mathbf{u}\in\mathbb{A}, \\
       \infty &\text{if}\quad \mathbf{u}\notin\mathbb{A},
       \end{cases}
   \end{align} 
   where
     \begin{align*}
       \mathcal{D}_{\boldsymbol{\phi}}(\mathbf{u,a,b})\coloneqq\sum_{k=1}^K\int_{\Omega}(a_k-\phi_k(x))^2 u_k(x) + (b_k-\phi_k{(x))}^2(1-u_k{(x)})dx, \quad
         \mathcal{Q}(\mathbf{u}) \coloneqq\sum_{k=1}^K\lvert u_k\rvert_{\text{TV}}
         \end{align*}
Here, $\mathcal{D}$ is a fidelity term and $\mathcal{Q}$ is a regulariser that enforces the $u_k$s to be piecewise constant.
For fixed $\mathbf{u}$ the infimum of $\mathcal{D}$ in~\eqref{energyred} is given by
\begin{align*}
    \boldsymbol{A}(\boldsymbol{u},\boldsymbol{\phi})\coloneqq\prod_{k=1}^K A(u_k,\phi_k)\times\prod_{k=1}^K A(1-u_k,\phi_k)
    \end{align*}
    where
     \begin{align*}
     A(u_k,\phi_k)\coloneqq\begin{cases}
     \mathbb{R} & \text{if}\quad u_k=0\\
     \frac{1}{\lVert u_k\rVert_1}\int_{\Omega}\phi_k(x) u_k(x)dx &\text{otherwise}.
     \end{cases}    
     \end{align*}
     Note that, if $u_k=0$ any $a_k\in\mathbb{R}$ is a minimiser of $\mathcal{D}_{\phi}(u,\cdot).$ 
The regularisation parameter for the total variation (TV) is given by $\lambda>0$.
 For the efficient minimisation of our energy functional in~\cite{gruber2022joint}, we implement a non-convex version of the first-order primal-dual algorithm proposed by Chambolle and Pock in~\cite{valkonen2014primal, chambolle2011first}. In this recently submitted paper~\cite{gruber2022joint}, we have shown that at least one global minimiser of \eqref{energyred} exists, as well as that the minimisation of the functional is stable with respect to perturbation of the image. We also showed the convergence to $\mathcal{Q}$-minimising solutions $\mathbf{u}$ with $\mathcal{R}_{\phi,\lambda}(\mathbf{u}^\ast)=0$ in the special case where the feature images are piece-wise constant.
 

\subsection{CNN Augmentation}
As already mentioned, a strength of the method~\cite{gruber2022joint} is the possibility to combine it with convolutional neural networks which we implement in this work. For this purpose, a CNN consisting of four convolutional blocks is trained, each of them made up of 32 channels with convolutional kernel size 3, which divides the image into the $K$ different feature maps. More specifically, we design a neural network, that consists of a decomposition path, that takes as input an image that has one channel. The decomposition path $\mathbf{\Psi}^{\theta}$ consists of three convolutional blocks, where the last one has three output channels. These intermediate outputs serve as the feature maps. In a subsequent layer, these outputs are summed up and a further convolutional block, $\tilde{\mathbf{\Psi}^\eta}$ is applied. The final (second) output of the network has one channel and represents the reconstruction of the original input $f$. As a loss function we use the MSE loss together with auxiliary losses on the intermediate outputs preventing empty components and enforcing pairwise uncorrelated feature maps~\cite{kiechle2018model} within the $K$ decompositions.
\noindent More specifically, let $\boldsymbol{\Psi}_k^{\theta}(f), k=1,\dots,K$ denote the extracted feature maps and by $\tilde{\boldsymbol{\Psi}}^{\eta}$ we denote the aforemmentioned reconstruction part of our network, i.e. $\tilde{\boldsymbol{\Psi}}^{\eta}$ takes as input the sum of the $K$ extracted feature maps, and outputs a reconstructed version of the image $f$. Then the loss function can be formulated as
\begin{align}\label{eq:loss}
    \mathcal{L}(\theta,\eta)=\alpha_1\sum_{k=1}^K\left(\lVert\boldsymbol{\Psi}_k^{\theta}(f)\rVert_2-\frac{1}{K}\lVert f\rVert_2\right)^2
     \nonumber &-\alpha_2\sum_{k\neq j}\log\left(1-\frac{\langle\boldsymbol{\Psi}^{\theta}_k(f), \boldsymbol{\Psi}^{\theta}_j(f)\rangle}{\lVert\boldsymbol{\Psi}_k^{\theta}(f)\rVert_2\cdot\lVert\boldsymbol{\Psi}_j^{\theta}(f)\lVert_2}\right)\\ 
    &\hspace{0.1\textwidth}+(1-\alpha_1-\alpha_2)\lVert\tilde{\boldsymbol{\Psi}}^{\eta}(\sum_{k=1}^K\boldsymbol{\Psi}_k^{\theta}(f))-f\lVert_2^2.
\end{align}
In~\eqref{eq:loss}, the first term ensures that non of the feature maps has norm zero. The second term guarantees pairwise incoherence between channels and the last term is the reconstruction loss. The resulting $K$ feature maps serve as the input channels of energy~\eqref{energyred}. Before the feature maps are passed into the energy functional, they are normalised so that they take on values in $[0,1]$.
\begin{figure}[htp]
  \includegraphics[width=.95\textwidth]{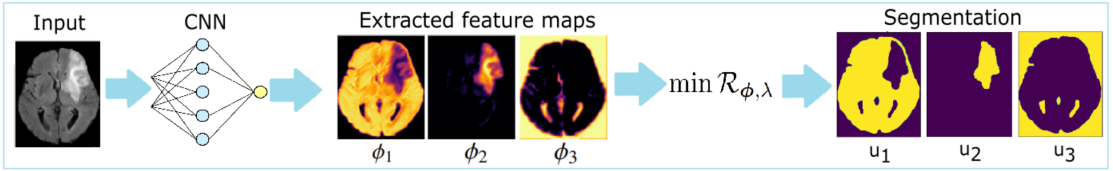}
  \vspace{-2mm}
\caption{The schematic representation of the proposed unsupervised segmentation workflow.}\label{texture}
 \label{fig:a}
\end{figure}

\section{RESULTS}\label{sec3}
In this section we demonstrate the performance of our approach and its capability of multiclass image segmentation of texture and medical images. For this purpose, we show the results obtained by minimisation of the energy functional~\eqref{energyred} after manual and CNN-based lifting. We present segmentations of a texture image, and another example, where we apply our CNN-augmented segmentation approach on a CT scan.

\subsection{Input lifting via Gabor filtering}
\noindent We consider a segmentation example based on texture information. For that purpose, we tune multiple Gabor filters with different spatial frequencies and orientations to capture texture. 
\begin{figure}[htp]
\centering
\includegraphics[width=.92\textwidth]{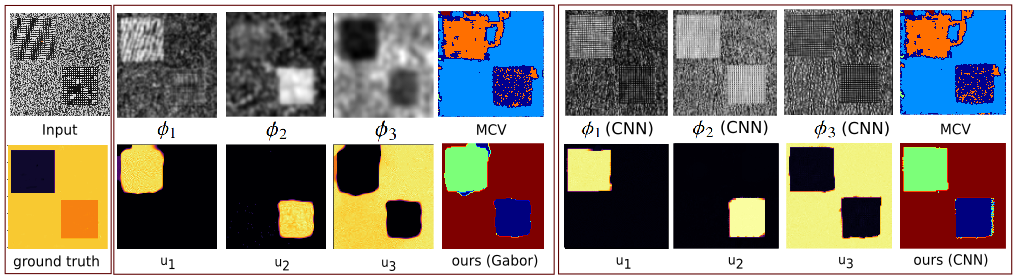}
\caption{Three-phase segmentation of a texture image using Gabor-based input lifting. The first box shows the input image and its corresponding ground truth. In the second box, the extracted feature maps using Gabor filters, and the results obtained by applying the multiphase Chan Vese (MCV) method ~\cite{vese2002multiphase} without pre-filtering, are illustrated. In the second row the ground truth segmentation as well as the minimisers of energy~\eqref{energyred} corresponding to the feature maps, are depicted. The third box shows the results obtained by applying the CNN-based lifting approach. }\label{brodatz}
\end{figure}
\noindent In Figure~\ref{brodatz} (left) we present results for the texture image, with its extracted feature maps (by using Gabor filters), and the corresponding segmentation masks which are the minimisers of~\eqref{energyred}. We compare our results with~\cite{vese2002multiphase}, where we use the MATLAB code provided by Y. Wu~\citep{alg} and set the regularisation parameter $\lambda$ to 0.5. For our method, the regularisation parameter $\lambda$ is set to 0.2. As can be seen from the images, the multiphase level set approach has difficulties on the segmentation of the texture image, while our approach does well. Note that the feature maps are obtained by summing filterings with Gabor kernels.
The parameters used for producing the pre-filterings are: $(\theta,\omega)\in\left\{(0,\frac{\sqrt{2}}{4}), (\frac{\pi}{4}, \frac{\sqrt{2}}{2})\right\}$ for $\phi_1$,  for the second feature map $\phi_2$ we have $(\theta,\omega)=(0,\frac{\sqrt{2}}{8})$, while 
for obtaining $\phi_3$ we use the sum of several Gabor filtered images with filter kernel parameters $(\theta,\omega)\in\left\{(0, \frac{\sqrt{2}}{32}),(0, \frac{\sqrt{2}}{16}),(\frac{\pi}{4}, \frac{\sqrt{2}}{64}),(\frac{\pi}{4}, \frac{\sqrt{2}}{32})\right\}$.
Using a CNN to obtain feature images allows us to skip the hand-crafted filter design and enables fully unsupervised segmentation.

\subsection{Segmentation using CNN-feature maps}\label{sec32}
\begin{figure}
\centering
  \includegraphics[width=.85\textwidth]{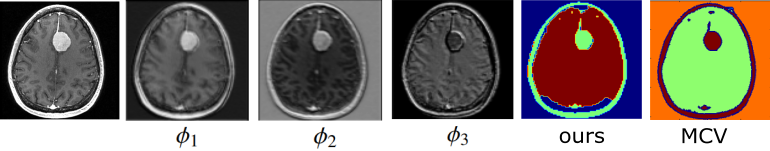}
\caption{Three-phase segmentation of a CT image of a patient with meningioma using the CNN-based input lifting and ~\cite{vese2002multiphase}.}\label{fig2}
\end{figure}

Here, we show the results obtained on a single-channel medical image. In Figure~\ref{brodatz}, the right-hand-side shows the results obtained by applying the CNN for input lifting and the obtained results. In Figure~\ref{fig2}, we apply our approach to a CT image showing a meningioma. To this end, we augment the input image by training the CNN described in section~\ref{sec2} to decompose a single image into $K$ differently filtered images. The example on the right in Figure~\ref{fig2} shows the CT image and corresponding extracted feature maps as well as the obtained segmentation masks using the proposed method and the level set approach from~\cite{vese2002multiphase}. We observe that in this example, the proposed method decomposes the image into three meaningful regions and can compete with the multiphase active contour model. 

\section{CONCLUSION AND OUTLOOK}
 
The basic idea of our method is to lift the input image to multichannel data using a CNN, which together with our previously proposed variational energy enables a fully unsupervised segmentation process. Numerical results show that for relatively simple examples, this approach leads to promising results. It has to be noted, that a the decomposition happens completely unsupervised/unguided, and therefore there is no guarantee that the results obtained follow expert knowledge.  To improve the unsupervised input lifting ensuring targeted image prefilterings alternative strategies and architectures will be explored in future work. One possible example is to use a patch-guided approach, which introduces further region-related information into decomposition process. 
Furthermore, we emphasise that recently explored methods that use CNNs for image representation can be used to decompose the images. Techniques that have shown impressive results in image representation like Deep Image Prior~\cite{ulyanov2018deep} could be used to parameterise the feature maps. In addition, variational autoencoders (VAEs) trained on medical images of the class of interest could be utilized to obtain interesting image decompositions.

\bibliographystyle{splncs04}
%
\bibliography{ref}

\end{document}